\documentclass[letterpaper]{article} 
\usepackage{aaai2026}  
\usepackage{times}  
\usepackage{helvet}  
\usepackage{courier}  
\usepackage[hyphens]{url}  
\usepackage{graphicx} 
\usepackage{natbib}  
\usepackage{caption} 
\usepackage{algorithm}
\usepackage{algorithmic}
\usepackage{booktabs}
\usepackage{amsmath}
\usepackage{amssymb}
\usepackage{multirow}
\usepackage{caption}
\usepackage{graphicx}
\usepackage{xcolor}

\usepackage{newfloat}
\usepackage{listings}
\DeclareCaptionStyle{ruled}{labelfont=normalfont,labelsep=colon,strut=off} 
\floatstyle{ruled}
\newfloat{listing}{tb}{lst}{}
\floatname{listing}{Listing}
\title{Learning to Explore: Policy-Guided Outlier Synthesis for Graph Out-of-Distribution Detection}
\author{
    Li Sun\textsuperscript{\rm 1,2}\thanks{Corresponding Author: Li Sun.}, Lanxu Yang\textsuperscript{\rm 1}, Jiayu Tian\textsuperscript{\rm 2}, Bowen Fang\textsuperscript{\rm 3}, Xiaoyan Yu\textsuperscript{\rm 4},\\
    Junda Ye\textsuperscript{\rm 2}, Peng Tang\textsuperscript{\rm 5}, Hao Peng\textsuperscript{\rm 6}, Philip S. Yu\textsuperscript{\rm 7}
}
\affiliations{
    \textsuperscript{\rm 1}North China Electric Power University\\
    \textsuperscript{\rm 2}Beijing University of Posts and Telecommunications\\
    \textsuperscript{\rm 3}Tsinghua University\\
    \textsuperscript{\rm 4}Beijing Institute of Technology\\
    \textsuperscript{\rm 5}Shandong University\\
    \textsuperscript{\rm 6}Beihang University \\
    \textsuperscript{\rm 7}University of Illinois Chicago\\
    
    \{ccesunli, yanglanxu24\}@ncepu.edu.cn,
    jytian04@outlook.com,
    psyu@uic.edu
}

\begin{document}

\maketitle

\begin{abstract}
Detecting Out-of-Distribution (OOD) graphs—those are drawn from a different distribution from the training data-is a critical task for ensuring the safety and reliability of Graph Neural Networks. The main challenge in unsupervised graph-level Out-of-Distribution detection lies in its common reliance on purely in-distribution (ID) data. This ID-only training paradigm leads to an incomplete characterization of the feature space, resulting in decision boundaries that lack the robustness needed to effectively separate ID from OOD samples. While incorporating synthesized outliers into the training process is a promising direction, existing generation methods are limited by their dependence on pre-defined, non-adaptive sampling heuristics (e.g., distance- or density-based). Such fixed strategies lack the flexibility to systematically explore the most informative OOD regions for refining decision boundaries. To overcome this limitation, we propose a novel Policy-Guided Outlier Synthesis (PGOS) framework that replaces static heuristics with a learned, adaptive exploration policy. PGOS trains a reinforcement learning agent to autonomously navigate low-density regions within a structured latent space, sampling representations that are maximally effective for regularizing the OOD decision boundary. These sampled points are then decoded into high-quality pseudo-OOD graphs to enhance the detector's robustness. Extensive experiments demonstrate the strong performance of our method, state-of-the-art results on multiple graph OOD and anomaly detection benchmarks.
\end{abstract}


\section{Introduction}
Graph Neural Networks (GNNs) have become a cornerstone for graph-level classification, achieving state-of-the-art results in critical domains like molecular science \cite{lee2025pre} and social network analysis \cite{ijcai25info_diff,tois25event,tkde23align,tweb23align,ijcai2025p365}. However, this success relies on the standard assumption that test data is drawn from the same distribution as the training data \cite{kipf2017semi, hamilton2017inductive, velivckovic2018graph,zhuo2025a, yang2025disentangled, zhuo2025dualformer}. In real-world applications, this closed-world assumption is often violated, as deployed models inevitably encounter inputs from novel or shifted distributions—commonly known as Out-of-Distribution (OOD) samples. When faced with these OOD graphs, a standard GNN can fail silently, producing erroneous predictions with high confidence. This vulnerability poses a significant risk to model reliability and safety, making the detection of OOD inputs a critical challenge for building trustworthy graph learning systems \cite{icdm23deepricci,www25gfm}.
\begin{figure}[t]
\centering
\includegraphics[width=1.0\columnwidth]{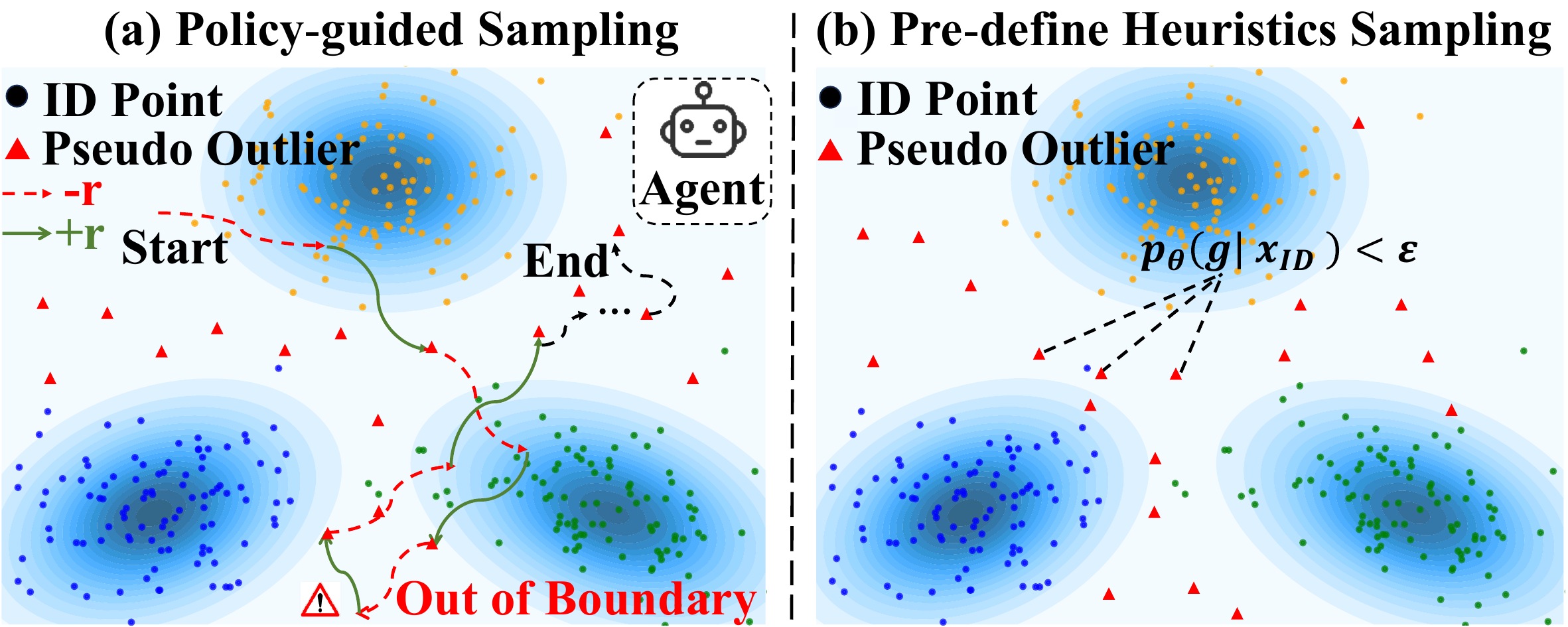} 
\caption{Comparisons between policy-guided and heuristic-based sampling. (a) Policy-guided sampling via a learned agent. (b) Heuristic sampling follows predefined rules.}
\label{intro}
\end{figure}

Existing research in unsupervised graph OOD detection has largely focused on modeling the in-distribution (ID) data manifold to indirectly identify outliers. Common strategies involve learning stable ID patterns through various means, such as contrastive learning \cite{liu2023good, hou2025structural}, generative modeling \cite{shen2024optimizing}, or post-hoc manipulations \cite{wang2024goodat, guo2023data}. A unified characteristic of these methods is their exclusive reliance on in-distribution data for training. When the data distribution is complex, relying solely on ID data may fail to reveal essential cues for OOD detection.

To overcome the limitations of ID-only training, an alternative paradigm is Outlier Exposure (OE) \cite{hendrycks2019oe} or Outlier Synthesis, which has been extensively validated in computer vision \cite{duvos, tao2023nonparametric, zheng2023out}. The core idea is to generate artificial outliers in the latent space to explicitly regularize the decision boundary. A common approach is to sample from low-likelihood regions of the in-distribution data, which can be modeled parametrically with class-conditional Gaussians \cite{duvos} or non-parametrically using k-NN density estimators \cite{tao2023nonparametric}. Other advanced strategies employ Hamiltonian Monte Carlo for more diverse exploration of the latent space \cite{lioutlier} or leverage powerful diffusion models to decode synthesized latent features into photo-realistic outlier images \cite{du2023dream}. However, the application of these powerful techniques to graph-structured data remains nascent, with HGOE representing a preliminary example that uses graph-based interpolation \cite{junwei2024hgoe}. A critical observation is that this entire diverse family of synthesis methods shares a fundamental limitation: they all rely on pre-defined heuristics—such as distance or density—to identify target regions for sampling. These fixed strategies lack the adaptability to systematically explore the most informative OOD regions for refining the decision boundaries. This raises a critical question: 

\textit{How can we move beyond fixed heuristics to systematically and adaptively discover the most informative outlier locations in the latent space?}

Tackling this question is particularly challenging in the unsupervised graph domain due to two fundamental issues. First, the latent space shaped by conventional contrastive learning is ill-suited for targeted exploration. While these methods can group similar graphs, they fail to establish explicit semantic prototypes for the clusters. This absence of learnable prototypes renders the low-density regions between clusters truly unstructured and difficult to navigate. Second, designing an adaptive exploration policy is itself a non-trivial task. To be effective, the policy must be guided by a carefully crafted reward mechanism that goes beyond simple metrics like distance or density. The core challenge lies in precisely defining what constitutes an informative outlier and translating this abstract notion into a concrete learning signal. Therefore, a principled framework should first impose a meaningful structure on the latent space and then learn a systematic policy for exploration.

To address this, we propose a novel framework Policy-Guided Outlier Synthesis (PGOS) that first imposes a meaningful structure on the latent space. Specifically, we employ prototypical contrastive learning \cite{PCL, lin2022prototypical} to establish the explicit learnable prototypes that conventional methods lack. These prototypes act as semantic anchors transforming the previously unstructured regions into a navigable space defined by well-separated clusters of ID data. Building upon this structured latent space, PGOS reframes outlier synthesis as a targeted exploration problem and tackles the policy design challenge by introducing a highly specialized RL agent. This agent’s behavior is governed by three key principles: (1) a tailored reward function that compels exploration into the low-density regions between prototypes; (2) a hard boundary constraint that ensures exploration remains relevant to the data manifold; (3) a novel spatially-aware entropy regularization method that dynamically encourages maximal exploration near the cluster boundaries. This principled guidance system enables the agent to learn an optimal policy for discovering latent representations that are maximally effective for regularizing the OOD decision boundary. These representations are then decoded into high-quality pseudo-OOD graphs to enhance robust model training. Our main contributions are as follows:
\begin{itemize}
    \item We reconsider outlier synthesis for graph OOD detection by adaptively exploring the ID-OOD boundary with a learnable policy beyond static heuristics.
    \item We design a policy-guided agent whose novel exploration strategy integrates a tailored reward, boundary constraints, and adaptive entropy regularization to efficiently discover informative pseudo-outliers.
    \item Extensive experiments on 25 benchmarks validate our method's superiority, where we establish new state-of-the-art performance on 12 of these datasets.
\end{itemize}
\section{Related Work}
\subsubsection{Graph-level Out-of-Distribution Detection.}
Unsupervised graph-level Out-of-Distribution detection aims to identify graphs that deviate from the training distribution. The current dominant paradigm is contrastive learning methods like GOOD-D \cite{liu2023good}, which captures hierarchical semantics, and SEGO \cite{hou2025structural}, which leverages structural entropy \cite{li2016structural}. Other approaches exploit intrinsic graph properties like substructures or spectral anomalies \cite{gu2025spectralgap, ding2024sgood}, rely on the reconstruction quality of generative models \cite{shen2024optimizing}, or explicitly model the class-conditioned distributions \cite{zhang2025enhancing}. Additionally, post-hoc methods such as AAGOD \cite{guo2023data} enhance pre-trained detectors, while test-time strategies like GOODAT \cite{wang2024goodat} perform detection without accessing training data. Orthogonal to these methods, other work provides statistical guarantees via conformal prediction \cite{lin2025conformal}. Recently, inspired by Outlier Exposure in computer vision, HGOE \cite{junwei2024hgoe} incorporates external real-world and internal synthetic outliers. However, its synthesis relies on a pre-defined interpolation heuristic. In contrast, our framework employs an adaptive policy to actively discover informative outliers beyond such fixed generation rules.
\subsubsection{Outlier Synthesis for OOD Detection.}
Outlier synthesis improves OOD detection by generating pseudo-samples that mimic distributional shifts. Representative methods like VOS \cite{duvos} and HamOS \cite{lioutlier} create boundary-aware outliers in feature space, while diffusion-based \cite{du2023dream}, flow-based \cite{kumar2023normalizing}, and non-parametric \cite{tao2023nonparametric} approaches enable fine-grained control over sample diversity and location. Recent advances explore diversity-aware sampling \cite{jiangdiverse, zhu2023diversified}, noisy outlier robustness \cite{zheng2023out}, and structural awareness via energy-based memory \cite{hofmann2024energy}.
Yet, most methods adopt predefined generation objectives, lacking adaptive exploration of the data manifold. This hinders their ability to generate semantically aligned and decision-relevant outliers.

\section{Preliminaries and Problem Definition}
An attributed graph is denoted as \( G = (\mathcal{V}, \mathcal{E}) \), where \( \mathcal{V} = \{v_1, \ldots, v_n\} \) is the set of \( n = |\mathcal{V}| \) nodes, and \( \mathcal{E} \subseteq \mathcal{V} \times \mathcal{V} \) is the set of edges. The topological structure of the graph can be described by an adjacency matrix \( A \in \{0, 1\}^{n \times n} \), where \( A_{ij} = 1 \) if there exists an edge between nodes \( v_i \) and \( v_j \), and \( A_{ij} = 0 \) otherwise. Each node \( v_i \in \mathcal{V} \) is associated with a \( d \)-dimensional feature vector. The features of all nodes are collectively represented by a node feature matrix \( X \in \mathbb{R}^{n \times d} \). Consequently, a graph can be concisely represented as the tuple \( G = (A, X) \).

\subsubsection{Unsupervised Graph-level OOD Detection.}
In this paper, we focus on the task of unsupervised graph-level out-of-distribution detection. In this setting, we are given a training set of unlabeled graphs, $\mathcal{D}_\text{train}^\text{in}=\{G_1, G_2,\ldots, G_n\}$, which are assumed to be sampled from an in-distribution, denoted as $\mathbb{P}_\text{in}$. The goal is to learn a scoring function, $f(\cdot)$, using only this ID training set. During the testing phase, this scoring function is used to assign an OOD score, $s = f(G)$, to any given graph $G$. A higher score indicates a greater likelihood that the graph originates from an unknown out-of-distribution, $\mathbb{P}_\text{out}$, where $\mathbb{P}_\text{in}\neq\mathbb{P}_\text{out}$. The model's performance is evaluated on a test set composed of unseen ID samples $\mathcal{D}_\text{test}^\text{in}$ and OOD samples $\mathcal{D}_\text{test}^\text{out}$, where $\mathcal{D}_\text{train}^\text{in}\cap\mathcal{D}_\text{test}^\text{in}=\varnothing$.
\subsubsection{Reinforcement Learning.}
Reinforcement Learning \cite{sutton1998reinforcement} provides a mathematical framework for an agent to learn an optimal policy $\pi$ by interacting with an environment, which is formally modeled as a Markov Decision Process (MDP). An MDP is defined by a state space $\mathcal{S}$, an action space $\mathcal{A}$, a transition function $P(s'|s, a)$, a reward function $R(s, a, s')$, and a discount factor $\gamma \in [0, 1)$.

The agent's goal is to learn a policy $\pi(a|s)$ that maximizes the expected discounted sum of future rewards. The value of taking an action $a$ in a state $s$ is given by the action-value function (Q-function):
\begin{equation}
    Q^\pi(s, a) = \mathbb{E}_{\pi} \left[ \sum_{k=0}^{\infty} \gamma^k r_{t+k+1} \middle| s_t = s, a_t = a \right]
\end{equation}

The objective of RL is to find the optimal Q-function $Q^*(s, a) = \max_{\pi} Q^\pi(s, a)$, which represents the maximum possible return. Once $Q^*(s, a)$ is determined, the optimal policy $\pi^*$ can be extracted by acting greedily at each state:
\begin{equation}
    \pi^*(s) = \arg\max_{a \in \mathcal{A}} Q^*(s, a)
\end{equation}

\begin{figure*}[t]
\centering
\includegraphics[width=1.0\textwidth]{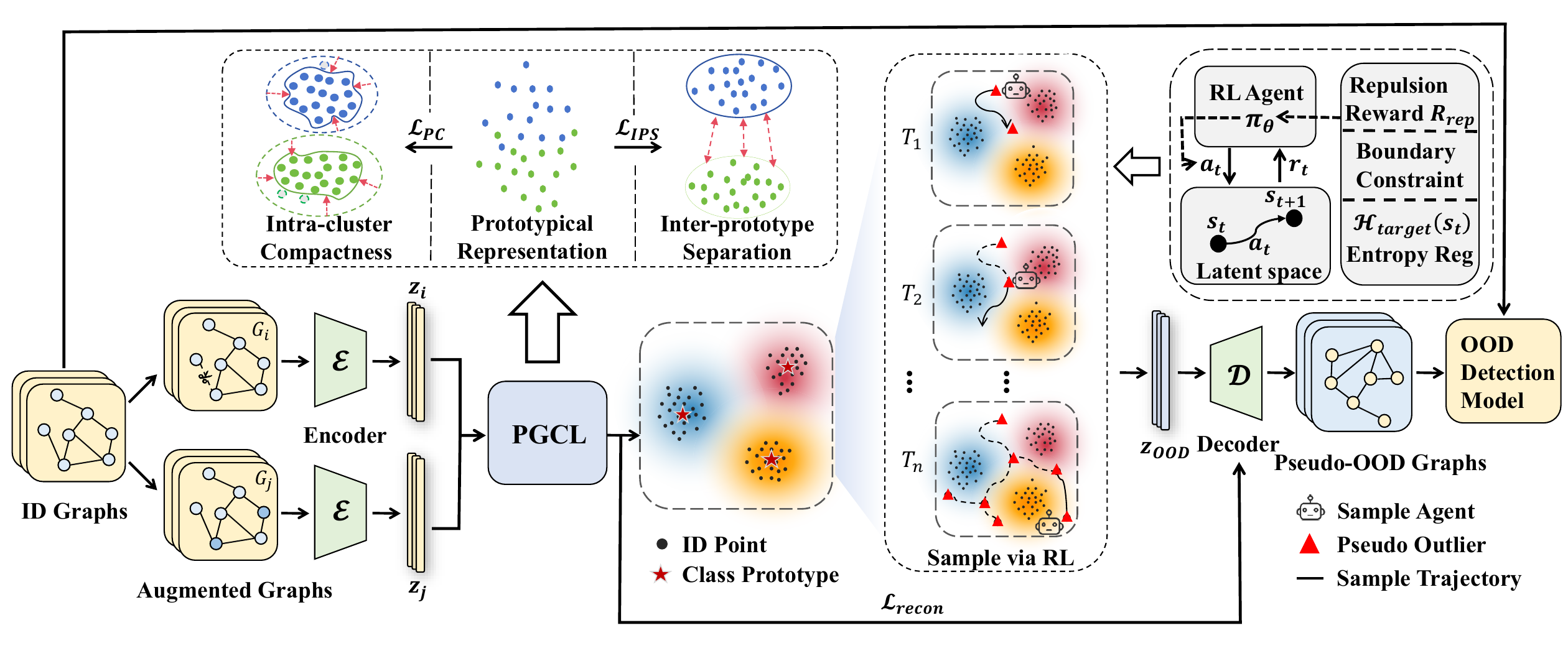} 
\caption{An overview of our PGOS framework based on pseudo-outlier synthesis.}
\label{framework}
\end{figure*}

\section{Methodology}
As illustrated in Figure \ref{framework}, we propose a pseudo-outlier synthesis framework for unsupervised graph OOD detection. We first train a graph autoencoder with prototypical contrastive learning to cluster semantically similar graphs around distinct prototypes. Then, a reinforcement learning-based sampler explores low-density regions in the latent space—far from any prototype—to generate informative latent vectors, which are decoded into high-quality pseudo-outlier graphs. Finally, we train the OOD detection model on both the synthesized outliers and the original ID graphs to distinguish between the two distributions. Technical details are provided in the following sections.                                                                                                        

\subsection{Prototypical Representation Learning for Graphs}
To effectively synthesize outliers, our strategy first requires a structured latent space that accurately models the in-distribution data. To this end, we design a prototypical representation learning module to produce a latent space where ID graphs form compact, well-separated clusters. This process yields two critical components: (1) a graph encoder $f_{\theta}$ that maps graphs into this structured space, consisting of a multi-layer GCN to learn node representations and a pooling layer to aggregate them into a graph-level embedding, and (2) a graph decoder $g_{\psi}$ that generates graph data from a given latent vector. The overall architecture for this module consists of the GCN-MLP encoder-decoder pair and a set of $K$ trainable prototypes $C = \{c_k\}_{k=1}^{K}$, where the number of prototypes $K$ is a pre-defined hyperparameter. We train the model by jointly optimizing a prototypical contrastive objective and a generative reconstruction loss.

\subsubsection{Prototypical Contrastive Objective.}
This objective $\mathcal{L}_\text{PCO}$ structures the latent space by pulling graph embeddings towards semantic prototypes and separating the prototypes themselves. By contrasting two augmented views of each graph $G_i$ with embeddings $z_i'$ and $z_i''$, this creates the low-density regions required for our sampling strategy. The objective is composed of three loss terms: debiased contrastive loss, prototypical consistency loss, and inter-prototype separation loss.

\textbf{Debiased Contrastive Loss} $\mathcal{L}_\text{DC}$ mitigates sampling bias of conventional contrastive learning by using prototype information to identify and exclude potential false negatives.
\begin{equation}
\mathcal{L}_\text{DC} = -\sum_{i=1}^{N} \log \frac{\text{sim}(z_i', z_i'')}{\text{sim}(z_i', z_i'') + \sum_{z_j \in \mathcal{N}_i} \text{sim}(z_i', z_j)}
\end{equation}
where we define the exponentiated similarity score as $\text{sim}(u, v) = \exp(u^T v / \tau)$, $\tau$ is the temperature, the set of debiased negative samples for an embedding $z_i'$ is $\mathcal{N}_i = \{z_j \mid j \neq i, c(z_j) \neq c(z_i')\}$, herein $c(z)$ denotes the nearest prototype for embedding $z$,

\textbf{Prototypical Consistency Loss} $\mathcal{L}_\text{PC}$ ensures clustering consistency between different augmented views of the same graph. It is defined as:
\begin{equation}
\mathcal{L}_\text{PC} = \frac{1}{2N} \sum_{i=1}^{N} \left[ l(z_i', z_i'') + l(z_i'', z_i') \right]
\end{equation}
where $N$ is the batch size and the loss term $l(z_a, z_b)$ is calculated by $l(z_a, z_b) = - \sum_{k=1}^{K} p_k(z_b) \log p_k(z_a)$. Here, $p_k(z) = \frac{\exp(z^T c_k / \tau)}{\sum_{j=1}^{K} \exp(z^T c_j / \tau)}$ is the predicted assignment probability of embedding $z$ to the $k^\text{th}$ prototype.

\textbf{Inter-Prototype Separation Loss} $\mathcal{L}_\text{IPS}$ is designed to push the prototypes away from each other. Its objective is to maximize the average squared Euclidean distance between all pairs of prototypes, thereby forcing them apart and creating well-defined, separated clusters.
\begin{equation}
\mathcal{L}_\text{IPS} = - \frac{1}{K(K-1)} \sum_{i=1}^{K} \sum_{j=1, j \ne i}^{K} ||c_i - c_j||_2^2
\end{equation}

These three components are combined into the prototypical contrastive objective as $\mathcal{L}_\text{PCO} = \mathcal{L}_\text{DC} + \mathcal{L}_\text{PC} + \mathcal{L}_\text{IPS}$.

\subsubsection{Generative Reconstruction Loss.}
To ensure the embeddings are informative and to train the decoder $g_{\psi}$, we employ a generative reconstruction loss $\mathcal{L}_\text{recon}$. The decoder aims to reconstruct both the graph's adjacency matrix $\hat{A}$ and its node features $\hat{X}$ from the latent node embeddings. The structure $\hat{A}$ is decoded via an inner product of embeddings, while an MLP decoder decodes the features $\hat{X}$. The reconstruction loss is defined as:
\begin{equation}
\mathcal{L}_\text{recon} = \sum_{i=1}^{N} \left( \|X_i - \hat{X}_i\|_{2}^{2} + \lambda \cdot \mathcal{L}_\text{struct}(A_i, \hat{A}_i) \right)
\end{equation}
where $\lambda$ is a balanced hyperparameter and $\mathcal{L}_\text{struct}$ is the element-wise Cross-Entropy loss for the adjacency matrices.

The final training objective consists of the prototypical contrastive objective and generative reconstruction losses:
\begin{equation}
\mathcal{L}_\text{total} = \mathcal{L}_\text{PCO} + \gamma\mathcal{L}_\text{recon}
\end{equation}
where $\gamma$ is a balancing hyperparameter. Minimizing the total objective shapes a structured latent space suitable for sampling and a decoder for generating pseudo-OOD graphs.

\subsection{Policy-Guided Outlier Synthesis}
Building upon a structured embedding space, we propose a reinforcement learning-based method that replaces previous non-adaptive sampling heuristics with an adaptive policy for exploratory outlier synthesis. Our approach introduces a novel guidance system for the RL agent, featuring a repulsion reward, hard boundary constraints, and dynamic spatial entropy regularization.
\subsubsection{MDP Formulation for Outlier Synthesis.}
We model the latent space $\mathcal{Z} \subset \mathbb{R}^D$ as an environment for the agent. The agent's goal is to learn an optimal policy $\pi$ for navigation and exploration. The agent's state $s_t$ is its current coordinate within the latent space. At each timestep, the agent takes an action $a_t$ which is a continuous displacement vector of the same dimension as the state. The subsequent state is determined by the transition $s_{t+1} = s_t + a_t$. The immediate reward $r_t$ is determined by the next state $s_{t+1}$ and is denoted by $r_t = R(s_{t+1})$. A high-quality OOD state should satisfy two key conditions: it must lie in the regions between the in-distribution clusters, and its position should not deviate excessively from the global boundary of these clusters.
\subsubsection{Reward Function Design.}
To guide the agent's exploration, our reward function is designed with a single and clear objective: to compel the agent to move into the low-density voids between in-distribution clusters. To achieve this, we introduce a repulsion reward $R_{\text{rep}}$ which penalizes the agent for entering the dense regions of the ID clusters. For each cluster $C_i$, we define its centroid $\mu_i$ and effective radius $r_i$ as $\mu_i = \frac{1}{|C_i|} \sum_{s \in C_i} s$ and $r_i = \max_{s \in C_i} \|s - \mu_i\|_2$. We apply a strong penalty when the agent's distance $d_i = \| s-\mu_i \|_2$ falls below a safety margin $\delta_i$, which is defined as a multiple of the cluster's specific radius $r_i$. As a result, the agent learns to focus its exploration on the spaces between clusters rather than the intra-cluster areas. The reward is formally expressed as:
\begin{equation}
    R_\text{rep}(s) = \sum_{i=1}^{K}
\begin{cases}
- \left(1 - \frac{d_i - r_i}{\delta_i}\right)^2 & \text{if }  d_i < r_i + \delta_i \\
0 & \text{otherwise}
\end{cases}
\end{equation}
\subsubsection{Boundary Constraint.}
To ensure the agent's exploration remains focused on areas relevant to the in-distribution data, we confine the state space $\mathcal{S}$ within a global boundary. This boundary is defined as a hypersphere centered at the global centroid of all ID embeddings, $\mu_g$, with a radius $R_\text{max}$. These parameters are pre-calculated from the training set $\mathcal{D}_\text{train}^\text{in}$ as $\mu_g = \frac{1}{|\mathcal{D}_\text{train}^\text{in}|} \sum_{s\in\mathcal{D}_\text{train}^\text{in}}s$ and $R_\text{max} = \max_{s \in \mathcal{D}_\text{train}^\text{in}} \|s - \mu_g\|_2$. Instead of using a reward penalty that can be inefficient for the agent to learn, we enforce a hard boundary constraint directly on the state transition dynamics. If an action $a_t$ causes the next state $s_{t+1} = s_t + a_t$ to lie outside this boundary, the agent's position is deterministically projected back onto the surface of the hypersphere.
\begin{equation}
s_{t+1} = \mu_g + R_\text{max} \frac{s_{t+1} - \mu_g}{\|s_{t+1} - \mu_g\|_2}
\end{equation}
This approach is more efficient and simplifies the agent's learning task.

\subsubsection{Spatially-Aware Entropy Regularization.}
To better explore the informative boundary regions of ID clusters, we innovate on the entropy regularization of Soft Actor-Critic (SAC) \cite{haarnoja2018soft}. Instead of using a manually designed entropy coefficient, we leverage SAC's automatic temperature tuning capability, guiding it with a dynamic target entropy. The core idea is to encourage maximal exploration near the boundaries of ID clusters. We first compute a characteristic length scale from data: the average radius of all ID clusters, $\bar{r} = \frac{1}{K} \sum_{i=1}^{K} r_i$. The target entropy $\mathcal{H}_\text{target}(s_t)$ for the agent's policy is then dynamically set based on the agent's distance to the nearest cluster, $d_{\min}(s_t)$, peaking when the agent is at the average boundary distance:
\begin{equation}
\mathcal{H}_\text{target}(s_t) = \mathcal{H}_\text{max} \cdot \exp\left(-\frac{(d_{\min}(s_t) - \bar{r})^2}{2\bar{r}^2}\right)
\end{equation}
where $\mathcal{H}_\text{max}$ is the maximum target entropy, often set to a default value based on the action space dimension. This self-tuning mechanism automatically focuses the agent's exploration on the most informative boundary regions.
\subsubsection{Learning the Exploration Policy.}
We employ the SAC algorithm to learn the exploration policy $\pi_{\phi}$. The architecture follows a standard actor-critic framework composed of a stochastic actor network $\pi_{\phi}$ to select actions and a pair of critic networks ($Q_{\theta_1}, Q_{\theta_2}$) to evaluate them. Both networks are trained concurrently using mini-batches of transitions sampled from a replay buffer. The actor's objective is to maximize both the expected return and the policy's entropy, where the entropy term encourages exploration. This is achieved by minimizing the following loss function:
\begin{equation}
\mathcal{L}_{\pi}(\phi) = \mathbb{E}_{ a_t \sim \pi_{\phi}} \left[ \alpha \log \pi_{\phi}(a_t|s_t) - Q_\text{min}(s_t, a_t) \right]
\end{equation}
where $a_t$ is an action sampled from the policy $\pi_{\phi}(\cdot|s_t)$, $Q_\text{min}(s_t, a_t)$ is the minimum estimated action-value from the pair of critic networks, i.e., $Q_\text{min}(s_t, a_t) = \min\limits_{i=1,2} Q_{\theta_i}(s_t, a_t)$ and $\alpha$ is the entropy temperature parameter. This parameter is automatically tuned towards a target $\mathcal{H}_{\text{target}}(s_t)$ by minimizing its own objective $J(\alpha)$. With a learning rate $\lambda_{\alpha}$, the update rule for $\alpha$ is:
\begin{equation}
\alpha \leftarrow \alpha - \lambda_\alpha \mathbb{E}_{a_t \sim\pi_{\phi}}\left[ -\log \pi_{\phi}(a_t|s_t)
- \mathcal{H}_{\text{target}}(s_t) \right]
\end{equation}The critic networks are trained concurrently by minimizing the standard soft Bellman error. 

Once the policy converges, we generate pseudo-OOD samples by executing the learned policy over multiple episodes to collect a set of latent representations. Each episode starts from the midpoint of two randomly selected prototype centroids. The pre-trained decoder then synthesizes these representations into final pseudo-OOD graphs until their total number matches that of the ID training set.

\subsection{Outlier-Regularized OOD Detection}
We instantiate our OOD detection model using GOOD-D \cite{liu2023good} and integrate the pseudo-outlier into the training process following HGOE \cite{junwei2024hgoe}. Specifically, we jointly optimize a standard OOD detection loss on in-distribution data and a boundary-aware regularization term on pseudo-outlier samples. The overall training objective is formulated as:
\begin{equation}
\mathcal{L} = \sum_{G \in \ D_\text{train}^\text{in}} \left[\mathcal{L}_{\text{ID}}(h(G)) + \beta \sum_{G' \in \ D_\text{OOD}}\mathcal{L}_\text{reg}(s_{G'})\right]
\end{equation}
where $\mathcal{L}_{\text{ID}}$ denotes the traning loss from GOOD-D, $ D_\text{OOD}$ is pseudo-OOD graphs sampling by our method, $s_{G'} = \text{sigmoid}(h(G'))$ is the OOD confidence score of a pseudo-outlier $G'$, and $\mathcal{L}_\text{reg}$ is the boundary-aware loss that penalizes uninformative outliers near or inside the in-distribution region and $\beta$ controls the regularization strength.

\section{Experiment}
\begin{table*}[t]
\centering
\setlength{\tabcolsep}{1mm}
\small
\begin{tabular}{l | cccccccccc|c}

\toprule
ID dataset & BZR & PTC-MR & AIDS & ENZYMES & IMDB-M & Tox21 & FreeSolv & BBBP & ClinTox & Esol & \multirow{2}{1.8em}{Avg. Rank}  \\
OOD dataset & COX2 & MUTAG & DHFR & PROTEIN & IMDB-B & SIDER & ToxCast & BACE & LIPO & MUV \\
\midrule
PK-SVM    & $43.1{\scriptstyle\pm7.9}$ & $50.2{\scriptstyle\pm6.7}$ & $52.1{\scriptstyle\pm2.6}$ & $49.9{\scriptstyle\pm3.2}$ & $49.0{\scriptstyle\pm3.0}$ & $51.0{\scriptstyle\pm2.0}$ & $49.3{\scriptstyle\pm2.9}$ & $53.0{\scriptstyle\pm2.3}$ & $51.9{\scriptstyle\pm3.0}$ & $51.7{\scriptstyle\pm1.8}$ &  $14.5$ \\
PK-iF & $52.8{\scriptstyle\pm2.2}$ & $52.0{\scriptstyle\pm4.0}$ & $50.8{\scriptstyle\pm1.8}$ & $52.6{\scriptstyle\pm2.4}$ & $51.9{\scriptstyle\pm3.0}$ & $50.7{\scriptstyle\pm1.1}$ & $51.9{\scriptstyle\pm1.7}$ & $53.1{\scriptstyle\pm1.8}$ & $51.7{\scriptstyle\pm1.6}$ & $51.2{\scriptstyle\pm4.1}$ &  $13.7$ \\
WL-SVM    & $50.2{\scriptstyle\pm5.2}$ & $52.4{\scriptstyle\pm7.8}$ & $51.4{\scriptstyle\pm3.3}$ & $52.9{\scriptstyle\pm3.2}$ & $52.9{\scriptstyle\pm2.6}$ & $50.7{\scriptstyle\pm1.9}$ & $50.4{\scriptstyle\pm3.8}$ & $52.9{\scriptstyle\pm2.0}$ & $50.8{\scriptstyle\pm3.7}$ & $52.0{\scriptstyle\pm2.2}$ &  $13.7$ \\
WL-iF   &   $50.0{\scriptstyle\pm3.0}$ & $52.8{\scriptstyle\pm1.9}$ & $50.8{\scriptstyle\pm0.9}$ & $51.4{\scriptstyle\pm3.0}$ & $52.6{\scriptstyle\pm3.5}$ & $51.3{\scriptstyle\pm0.6}$ & $53.0{\scriptstyle\pm3.2}$ & $50.6{\scriptstyle\pm0.7}$ & $51.4{\scriptstyle\pm2.5}$ & $51.6{\scriptstyle\pm1.3}$ & $14.0$\\

\midrule
IG-iF & $62.7{\scriptstyle\pm8.6}$ & $50.2{\scriptstyle\pm6.4}$ & $91.0{\scriptstyle\pm1.3}$ & $57.3{\scriptstyle\pm2.4}$ & $60.7{\scriptstyle\pm2.0}$ & $57.6{\scriptstyle\pm0.9}$ & $59.6{\scriptstyle\pm2.1}$ & $55.7{\scriptstyle\pm2.1}$ & $46.8{\scriptstyle\pm2.5}$ & $55.8{\scriptstyle\pm4.2}$ & $11.9$ \\
IG-MD & $85.8{\scriptstyle\pm6.3}$ & $52.6{\scriptstyle\pm8.2}$ & $70.8{\scriptstyle\pm11.3}$ & $53.2{\scriptstyle\pm4.2}$ & $80.9{\scriptstyle\pm0.5}$ & $58.6{\scriptstyle\pm1.2}$ & $59.1{\scriptstyle\pm5.7}$ & $72.3{\scriptstyle\pm4.3}$ & $46.9{\scriptstyle\pm4.7}$ & $75.6{\scriptstyle\pm2.5}$ & $9.9$ \\
GCL-iF & $58.9{\scriptstyle\pm5.4}$ & $52.4{\scriptstyle\pm2.7}$ & $91.3{\scriptstyle\pm2.1}$ & $60.8{\scriptstyle\pm3.1}$ & $58.0{\scriptstyle\pm2.5}$ & $58.0{\scriptstyle\pm1.4}$ & $56.7{\scriptstyle\pm1.5}$ & $57.9{\scriptstyle\pm2.0}$ & $48.8{\scriptstyle\pm1.4}$ & $60.3{\scriptstyle\pm3.7}$ & $11.2$ \\
GCL-MD & $84.8{\scriptstyle\pm5.4}$ & $70.9{\scriptstyle\pm2.0}$ & $92.0{\scriptstyle\pm1.0}$ & $51.9{\scriptstyle\pm4.4}$ & $81.2{\scriptstyle\pm1.6}$ & $60.7{\scriptstyle\pm1.3}$ & $59.7{\scriptstyle\pm4.3}$ & ${74.3{\scriptstyle\pm1.4}}$ & $53.3{\scriptstyle\pm2.6}$ & ${76.8{\scriptstyle\pm2.7}}$ & $8.5$ \\

\midrule
OCGIN  & $83.1{\scriptstyle\pm0.1}$ & ${74.0{\scriptstyle\pm0.6}}$ & $95.9{\scriptstyle\pm0.0}$ & $63.5{\scriptstyle\pm0.1}$ & $80.8{\scriptstyle\pm0.0}$ & $68.1{\scriptstyle\pm0.2}$ & $68.3{\scriptstyle\pm0.2}$ & $77.2{\scriptstyle\pm0.0}$ & $62.2{\scriptstyle\pm0.0}$ & $87.3{\scriptstyle\pm0.1}$ & $5.7$ \\
OCGTL  & $80.9{\scriptstyle\pm0.2}$ & $75.3{\scriptstyle\pm0.6}$ & $\mathbf{99.3{\scriptstyle\pm0.0}}$ & $\underline{67.6{\scriptstyle\pm0.0}}$ & $67.4{\scriptstyle\pm0.7}$ & $60.9{\scriptstyle\pm0.0}$ & $64.6{\scriptstyle\pm0.0}$ & $76.3{\scriptstyle\pm0.0}$ & $54.8{\scriptstyle\pm1.1}$ & $84.9{\scriptstyle\pm0.0}$ & $6.3$ \\
SIGNET & ${86.9{\scriptstyle\pm0.3}}$ & $\underline{81.9{\scriptstyle\pm0.6}}$ & $97.3{\scriptstyle\pm0.0}$ & $62.1{\scriptstyle\pm0.1}$ & $\underline{83.0{\scriptstyle\pm0.1}}$ & ${65.6{\scriptstyle\pm0.1}}$ & $\mathbf{75.4{\scriptstyle\pm0.1}}$ & $\mathbf{91.4{\scriptstyle\pm0.0}}$ & $\underline{73.0{\scriptstyle\pm0.1}}$ & $87.2{\scriptstyle\pm0.0}$ & $3.4$ \\
CVTGAD & $\mathbf{96.9}{\mathbf{\scriptstyle\pm0.0}}$ & $80.3{\scriptstyle\pm0.3}$ & $\underline{99.0{\scriptstyle\pm0.0}}$ & $65.4{\scriptstyle\pm0.1}$ & $82.5{\scriptstyle\pm0.0}$ & $\underline{68.1{\scriptstyle\pm0.1}}$ & $73.0{\scriptstyle\pm0.2}$ & $87.1{\scriptstyle\pm0.0}$ & $67.8{\scriptstyle\pm0.0}$ & $\mathbf{92.7}{\mathbf{\scriptstyle\pm0.0}}$ & $\underline{2.6}$ \\
GLocalKD  & $80.7{\scriptstyle\pm0.2}$ & $75.0{\scriptstyle\pm0.6}$ & $94.0{\scriptstyle\pm0.0}$ & $58.9{\scriptstyle\pm0.1}$ & $81.4{\scriptstyle\pm0.1}$ & $56.7{\scriptstyle\pm0.1}$ & $70.7{\scriptstyle\pm0.3}$ & $68.0{\scriptstyle\pm0.4}$ & $56.7{\scriptstyle\pm0.1}$ & $88.0{\scriptstyle\pm0.3}$ & $7.0$ \\
\midrule
GOOD-D & $75.8{\scriptstyle\pm6.0}$ & $70.6{\scriptstyle\pm3.5}$ & $93.7{\scriptstyle\pm1.2}$ & $57.2{\scriptstyle\pm2.0}$ & $78.2{\scriptstyle\pm4.3}$ & $66.3{\scriptstyle\pm1.0}$ & $64.8{\scriptstyle\pm3.3}$ & $73.2{\scriptstyle\pm1.3}$ & $55.7{\scriptstyle\pm3.8}$ & $86.8{\scriptstyle\pm2.4}$ & $8.0$ \\
GOODAT & $\underline{96.5{\scriptstyle\pm0.0}}$ & ${81.2{\scriptstyle\pm0.2}}$ & $98.6{\scriptstyle\pm0.0}$ & $65.1{\scriptstyle\pm0.0}$ & $81.0{\scriptstyle\pm0.1}$ & $66.8{\scriptstyle\pm0.0}$ & $68.5{\scriptstyle\pm0.3}$ & $83.1{\scriptstyle\pm0.1}$ & $68.5{\scriptstyle\pm0.0}$ & ${89.6{\scriptstyle\pm0.0}}$ & $3.7$ \\
\midrule
PGOS & $94.8{\scriptstyle\pm1.1}$ & $\mathbf{84.1}{\mathbf{\scriptstyle\pm1.7}}$ & ${96.8}{\scriptstyle\pm2.3}$ & $\mathbf{68.9}{\mathbf{\scriptstyle\pm1.6}}$ & $\mathbf{85.4{\scriptstyle\pm1.4}}$ & $\mathbf{74.2}{\mathbf{\scriptstyle\pm1.1}}$ & $\underline{74.9{\mathbf{\scriptstyle\pm0.3}}}$ & $\underline{87.4{\scriptstyle\pm1.1}}$ & $\mathbf{75.2}{\mathbf{\scriptstyle\pm1.1}}$ & $ \underline{92.5{\scriptstyle\pm0.4}} $ & $\mathbf{1.9}$ \\
\bottomrule
\end{tabular}
\caption{OOD detection performance in AUC (\%, mean ± std), with the best and second-best results in \textbf{bold} and underlined.}
\label{tab:main_result_od}
\end{table*}
\subsection{Experimental Setup}
\subsubsection{Datasets.}
To ensure fair and consistent evaluation, we adopt the benchmark protocol introduced by \cite{liu2023good}, which covers tasks of graph OOD detection and graph anomaly detection. Besides, 15 datasets from the Tox21 challenge and the TU benchmark \cite{morris2020tudataset} are used for graph anomaly detection. Samples belonging to the minority class or the true anomalous class are treated as anomalies, whereas all others are considered normal. For all datasets, we adopt the same splitting strategy as defined in the UB-GOLD benchmark \cite{wang2025unifying}. 

\subsubsection{Baselines.}
 We compare the baselines across four main categories, including 4 graph kernel with detector methods \cite{li2016detecting, neumann2016propagation,liu2008isolation,amer2013enhancing}, 4 SSL with detector methods  \cite{sun2020infograph,you2020graph,sehwagssd}, 5 GNN-based graph-level anomaly detection methods like OCGTL \cite{qiu2022ocgtl}, SIGNET \cite{liu2023signet}, OCGIN \cite{zhao2023ocgin}, CVTGAD  \cite{li2023cvtgad}, GLocalKD \cite{ma2023glocalkd} and 2 GNN-based graph-level out-of-distribution detection methods  \cite{liu2023good,wang2024goodat}.

\subsubsection{Evaluation \& Implementation.}
We evaluate OOD detection performance using the AUC metric. We report the mean and standard deviation across 5 independent runs. Complete implementation details, hyperparameter settings, and additional experimental results are provided in the appendix.

\subsection{Experimental Results and Analysis}
\begin{figure}[t]
\centering
\includegraphics[width=1.0\columnwidth]{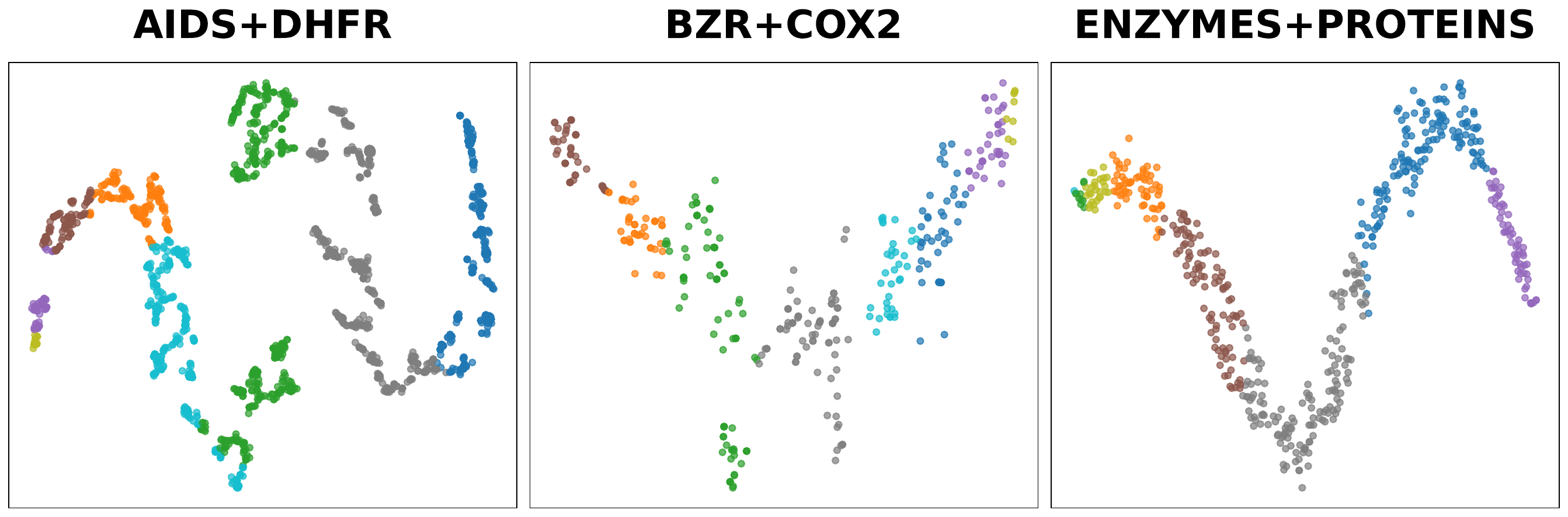} 
\caption{T-SNE visualizations of graph embeddings learned by PGCL on three different datasets. Each point denotes a graph, and colors represent the $K=8$ distinct clusters.}
\label{emb_visulization}
\end{figure}
To evaluate the effectiveness of our proposed method, we conduct comprehensive experiments including: (1) performance evaluation on OOD detection and anomaly detection tasks, (2) ablation study, (3) visualization analysis, and (4) sensitivity analysis of the number of prototypes. 

\subsubsection{Performance Evaluation on OOD Detection.}
As shown in Table \ref{tab:main_result_od}, we compare our proposed PGOS method with 15 competitive baselines across 10 representative OOD detection benchmarks. PGOS achieves the best average rank of 1.9, demonstrating strong overall performance and generalization across diverse distributional shifts. The performance gains are particularly obvious on several datasets. For instance, PGOS surpasses the second-best method by 2.2\% AUC on PTC-MR/MUTAG. On other benchmarks such as IMDB-M/IMDB-B and Tox21/SIDER, it also achieves significant improvements of 2.4\% and 6.1\% over the runner-up, respectively. Furthermore, even on challenging datasets like ENZYMES/PROTEIN where many methods struggle, PGOS achieves superior performance. These results consistently validate the effectiveness and robustness of our policy-guided outlier synthesis framework.

\subsubsection{Performance Evaluation on Anomaly Detection.}
\begin{table*}[ht]
\centering
\setlength{\tabcolsep}{1mm}
\small
\begin{tabular}{l | ccc|ccccc|cc|c}
\toprule
Method &  WL-SVM & IG-iF & GCL-iF & OCGIN & OCGTL & SIGNET & CVTGAD & GLocalKD & GOOD-D & GOODAT & PGOS \\
\midrule
PROTEINS &  $52.9{\scriptstyle\pm3.8}$ & $55.8{\scriptstyle\pm2.0}$ & $59.1{\scriptstyle\pm2.5}$ & $76.3{\scriptstyle\pm0.0}$ & $74.2{\scriptstyle\pm0.0}$ & $72.6{\scriptstyle\pm0.1}$ & $74.4{\scriptstyle\pm0.0}$ & $71.4{\scriptstyle\pm0.5}$ & $72.4{\scriptstyle\pm0.0}$ & $\underline{77.9{\scriptstyle\pm2.4}}$ & $\mathbf{81.3}{\mathbf{\scriptstyle\pm0.3}}$\\
ENZYMES  & $57.2{\scriptstyle\pm2.6}$ & $52.4{\scriptstyle\pm4.5}$ & $52.1{\scriptstyle\pm4.0}$ & $63.6{\scriptstyle\pm2.2}$ & $66.0{\scriptstyle\pm0.1}$ & $63.2{\scriptstyle\pm0.3}$ & $\mathbf{69.4{\scriptstyle\pm0.7}}$ & $66.0{\scriptstyle\pm0.4}$ & $61.5{\scriptstyle\pm0.7}$ & $59.1{\scriptstyle\pm2.5}$ & $\underline{67.1{\scriptstyle\pm0.1}}$\\
AIDS & $49.2{\scriptstyle\pm3.4}$ & $72.8{\scriptstyle\pm2.4}$ & $78.0{\scriptstyle\pm2.7}$ & $\mathbf{99.8}{\mathbf{\scriptstyle\pm0.0}}$ & $\underline{99.6{\scriptstyle\pm0.0}}$ & $92.2{\scriptstyle\pm0.3}$ & $98.6{\scriptstyle\pm0.0}$ & $97.3{\scriptstyle\pm0.0}$ & $95.9{\scriptstyle\pm0.1}$ & $96.0{\scriptstyle\pm1.0}$ & $96.9{\scriptstyle\pm0.1}$\\
DHFR  & $52.9{\scriptstyle\pm3.0}$ & $51.4{\scriptstyle\pm3.0}$ & $50.3{\scriptstyle\pm2.8}$ & $59.8{\scriptstyle\pm0.2}$ & $59.4{\scriptstyle\pm0.1}$ & $\underline{72.7{\scriptstyle\pm0.6}}$ & $64.2{\scriptstyle\pm0.1}$ & $59.6{\scriptstyle\pm0.1}$ & $64.5{\scriptstyle\pm0.0}$ & $62.2{\scriptstyle\pm2.5}$ & $\mathbf{73.2}{\mathbf{\scriptstyle\pm0.2}}$\\
BZR & $52.3{\scriptstyle\pm6.2}$ & $61.2{\scriptstyle\pm2.5}$ & $61.2{\scriptstyle\pm2.8}$ & $68.9{\scriptstyle\pm0.4}$ & $63.5{\scriptstyle\pm0.6}$ & $\underline{79.3{\scriptstyle\pm1.3}}$ & $\mathbf{80.0}{\mathbf{\scriptstyle\pm0.5}}$ & $63.9{\scriptstyle\pm1.3}$ & $65.7{\scriptstyle\pm0.6}$ & $59.0{\scriptstyle\pm5.3}$ & $77.1{\scriptstyle\pm2.4}$\\
COX2  & $49.2{\scriptstyle\pm2.9}$ & $55.7{\scriptstyle\pm2.9}$ & $51.0{\scriptstyle\pm1.7}$ & $57.4{\scriptstyle\pm0.5}$ & $56.6{\scriptstyle\pm0.3}$ & $\underline{70.6{\scriptstyle\pm0.2}}$ & $67.6{\scriptstyle\pm0.4}$ & $49.2{\scriptstyle\pm0.6}$ & $61.6{\scriptstyle\pm0.7}$ & $59.3{\scriptstyle\pm7.6}$ & $\mathbf{74.6}{\mathbf{\scriptstyle\pm2.3}}$\\
DD & $46.2{\scriptstyle\pm1.1}$ & $57.8{\scriptstyle\pm2.8}$ & $53.8{\scriptstyle\pm2.0}$ & $\underline{80.5{\scriptstyle\pm0.0}}$ & $80.4{\scriptstyle\pm0.0}$ & $74.1{\scriptstyle\pm0.1}$ & $74.2{\scriptstyle\pm0.0}$ & $78.3{\scriptstyle\pm1.4}$ & $78.5{\scriptstyle\pm0.0}$ & $78.2{\scriptstyle\pm2.2}$ & $\mathbf{81.1}{\mathbf{\scriptstyle\pm0.4}}$\\
NCI1 & $52.9{\scriptstyle\pm1.4}$ & $52.4{\scriptstyle\pm1.9}$ & $50.7{\scriptstyle\pm1.7}$ & $52.2{\scriptstyle\pm0.1}$ & $\mathbf{79.0{\scriptstyle\pm0.0}}$ & $69.1{\scriptstyle\pm0.0}$ & $75.0{\scriptstyle\pm0.0}$ & $61.1{\scriptstyle\pm0.1}$ & $60.9{\scriptstyle\pm0.1}$ & $45.6{\scriptstyle\pm1.0}$ & $\underline{{76.1}{\mathbf{\scriptstyle\pm0.1}}}$\\
IMDB-B  & $52.7{\scriptstyle\pm1.8}$ & $64.9{\scriptstyle\pm3.1}$ & $54.9{\scriptstyle\pm3.5}$ & $61.2{\scriptstyle\pm0.2}$ & $63.1{\scriptstyle\pm0.0}$ & $\underline{70.3{\scriptstyle\pm0.1}}$ & $\mathbf{71.5{\scriptstyle\pm0.2}}$ & $50.6{\scriptstyle\pm0.2}$ & $67.6{\scriptstyle\pm0.2}$ & $65.5{\scriptstyle\pm4.3}$ & ${67.7}{{\scriptstyle\pm2.4}}$\\
REDDIT-B & $47.4{\scriptstyle\pm2.9}$ & $67.5{\scriptstyle\pm3.0}$ & $75.0{\scriptstyle\pm1.6}$ & $\mathbf{93.3}{\mathbf{\scriptstyle\pm0.0}}$ & $\underline{89.7{\scriptstyle\pm0.0}}$ & $85.8{\scriptstyle\pm0.1}$ & $87.7{\scriptstyle\pm0.0}$ & $79.5{\scriptstyle\pm0.1}$ & $88.5{\scriptstyle\pm0.0}$ & $80.3{\scriptstyle\pm0.8}$ & $86.4{\scriptstyle\pm0.7}$\\
COLLAB  & $62.0{\scriptstyle\pm1.8}$ & $45.0{\scriptstyle\pm2.8}$ & $49.3{\scriptstyle\pm2.8}$ & $62.2{\scriptstyle\pm0.1}$ & $51.4{\scriptstyle\pm0.0}$ & $\underline{71.2{\scriptstyle\pm0.1}}$ & $65.2{\scriptstyle\pm0.0}$ & $48.0{\scriptstyle\pm0.0}$ & $63.4{\scriptstyle\pm0.0}$ & $45.3{\scriptstyle\pm0.9}$ & $\mathbf{74.5}{\mathbf{\scriptstyle\pm1.2}}$\\
HSE & $64.9{\scriptstyle\pm1.1}$ & $51.5{\scriptstyle\pm3.0}$ & $50.9{\scriptstyle\pm1.8}$ & ${72.3{\scriptstyle\pm0.1}}$ & $59.2{\scriptstyle\pm0.0}$ & $67.2{\scriptstyle\pm0.0}$ & $67.0{\scriptstyle\pm0.3}$ & $58.8{\scriptstyle\pm0.0}$ & $\underline{72.7{\scriptstyle\pm0.1}}$ & $63.4{\scriptstyle\pm1.1}$ & $\mathbf{78.6}{\mathbf{\scriptstyle\pm0.0}}$\\
MMP  & $58.3{\scriptstyle\pm2.0}$ & $56.9{\scriptstyle\pm1.8}$ & $54.3{\scriptstyle\pm1.6}$ & $69.0{\scriptstyle\pm0.0}$ & $60.8{\scriptstyle\pm0.5}$ & $\mathbf{75.1{\scriptstyle\pm0.1}}$ & $69.9{\scriptstyle\pm0.0}$ & $60.8{\scriptstyle\pm0.4}$ & $71.6{\scriptstyle\pm0.0}$ & $69.4{\scriptstyle\pm0.4}$ & $\underline{{71.7}{\scriptstyle\pm0.1}}$\\
p53  & $59.8{\scriptstyle\pm2.8}$ & $52.1{\scriptstyle\pm1.6}$ & $55.0{\scriptstyle\pm2.1}$ & $\underline{69.3{\scriptstyle\pm0.1}}$ & $66.4{\scriptstyle\pm0.1}$ & $66.4{\scriptstyle\pm0.1}$ & $67.9{\scriptstyle\pm0.0}$ & $63.7{\scriptstyle\pm0.0}$ & ${68.7{\scriptstyle\pm0.0}}$ & $63.2{\scriptstyle\pm0.1}$ & $\mathbf{69.8}{\mathbf{\scriptstyle\pm0.1}}$\\
PPAR  & $58.9{\scriptstyle\pm1.8}$ & $53.4{\scriptstyle\pm2.9}$ & $51.8{\scriptstyle\pm1.8}$ & $65.0{\scriptstyle\pm0.0}$ & $64.7{\scriptstyle\pm0.0}$ & $\underline{{70.4\scriptstyle\pm0.2}}$ & $66.1{\scriptstyle\pm0.0}$ & $65.7{\scriptstyle\pm0.0}$ & $\mathbf{70.8{\scriptstyle\pm0.0}}$ & $66.8{\scriptstyle\pm2.5}$ & $68.0{\scriptstyle\pm0.1}$\\
\midrule
Avg. Rank  &$9.6$ & $9.8$ & $10.2$ & $4.5$ & $5.6$ & $4.1$ & $\underline{3.5}$ & $7.4$ & $4.2$ & $6.7$ & $\mathbf{2.1}$ \\
\bottomrule
\end{tabular}
\caption{Anomaly detection performance in AUC (\%, mean ± std), with the best and second-best results in bold and underlined.} 
\label{tab:main_result_ad}
\end{table*}

\begin{table}[htbp]
\centering
\small
\setlength{\tabcolsep}{1mm}
\begin{tabular}{c|cccc}
\toprule
\multirow{2}{*}{Model Variants} & BZR & PTC-MR & IMDB-M & Tox21 \\
 & COX2 & MUTAG & IMDB-B & SIDER \\
\midrule
PGOS-$\mathcal{L}_\text{IPS}$ & $93.3{\scriptstyle\pm1.1}$ & $82.2{\scriptstyle\pm2.2}$ & $83.7{\scriptstyle\pm1.4}$ & $72.9{\scriptstyle\pm1.7}$ \\
PGOS-PGCL& $93.4{\scriptstyle\pm2.7}$ & $80.9{\scriptstyle\pm3.7}$ & $81
.6{\scriptstyle\pm1.3}$ & $71.7{\scriptstyle\pm2.4}$ \\
PGOS-$\mathcal{H}_\text{target}(s_t)$& $93.1{\scriptstyle\pm1.3} $ & $82.8{\scriptstyle\pm2.1}$& $82.0{\scriptstyle\pm1.3}$ & $70.6{\scriptstyle\pm1.4} $\\
PGOS-RL& $77.5{\scriptstyle\pm1.6} $ & $74.5{\scriptstyle\pm1.3} $& $76.8{\scriptstyle\pm2.6}$ & $64.9{\scriptstyle\pm2.3} $\\
\midrule
PGOS-Full & $\mathbf{94.8}\mathbf{{\scriptstyle\pm1.1}} $ & $\mathbf{84.1}\mathbf{{\scriptstyle\pm1.7}} $ & 
$ \mathbf{85.4}\mathbf{{\scriptstyle\pm1.4}}$ & $\mathbf{74.2}\mathbf{{\scriptstyle\pm1.1}}$ \\
\hline
\end{tabular}
\caption{Performance comparison of PGOS and its ablated variants, reported as AUC scores (\%, mean ± std).}
\label{ablation}
\end{table}

\begin{figure}[t]
\centering
\includegraphics[width=1.0\columnwidth]{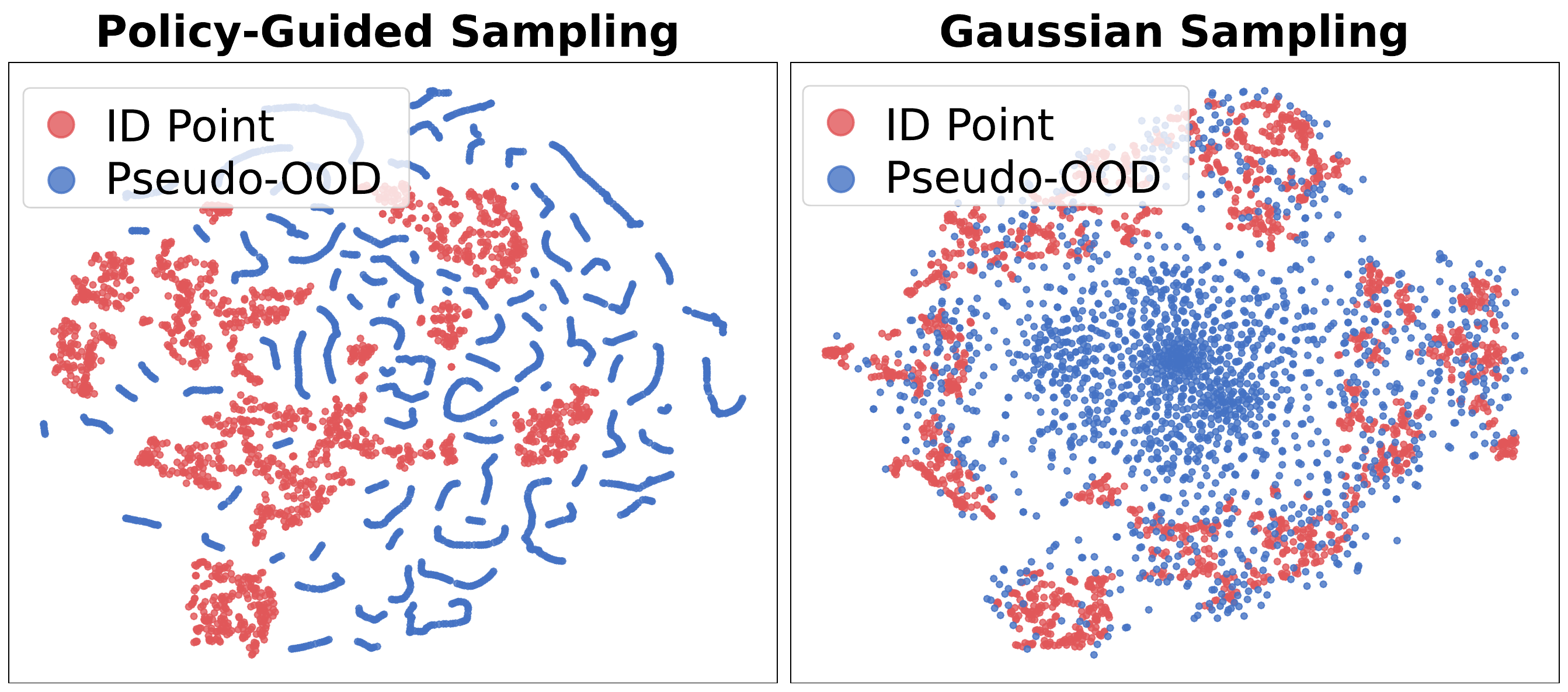} 
\caption{T-SNE visualization of sampled points generated by the RL-based and Gaussian sampling strategies on the BBBP+BACE OOD dataset.}
\label{sampling}
\end{figure}

To evaluate the effectiveness of our proposed method in the anomaly detection setting, we conduct comprehensive experiments on 15 graph-level anomaly detection benchmark datasets. As shown in Table \ref{tab:main_result_ad}, our method achieves new state-of-the-art performance on 7 out of 15 datasets. Even on datasets where PGOS is not the leading method, its performance remains highly competitive. Furthermore, PGOS demonstrates its superior performance most clearly on challenging datasets where other baselines struggle. For example, it achieves state-of-the-art performance on both HSE and COX2, outperforming the runner-up methods by significant margins of 5.9\% and 4.0\% AUC, respectively. These results confirm that our adaptive synthesis policy is a powerful and versatile solution for identifying both anomalous and out-of-distribution graphs.

\subsubsection{Ablation Study.}
As presented in Table \ref{ablation}, we analyze the crucial role of each component within the PGOS framework. The results demonstrate that the policy-guided sampling module is the most critical element; its removal (PGOS-RL) leads to a significant performance degradation across all datasets, causing the AUC score to drop by an average of 11.2\% and by as much as 17.3\% on the BZR/COX2 dataset.  This underscores the fundamental importance of an adaptive exploration policy. Furthermore, the other components also prove important for achieving optimal performance. Removing the inter-prototype separation loss (PGOS-$\mathcal{L}_{\text{IPS}}$) and the spatially-aware entropy regularization (PGOS-$\mathcal{H}_{\text{target}}(s_t)$) also results in noticeable performance drops, confirming their respective roles in structuring the latent space and efficiently guiding exploration. The superior performance of the full PGOS model over all ablated variants confirms that all components contribute synergistically to the framework's overall effectiveness.

\subsubsection{Visualization.}
To provide an intuitive understanding of the effectiveness of our PGCL module and policy-guided sampling strategy, we visualize both the learned embeddings and the sampled points on several representative datasets using T-SNE. As illustrated in Figure \ref{emb_visulization}, PGCL produces compact and well-separated clusters, effectively structuring the latent space across diverse dataset distributions and facilitating pseudo-outlier synthesis. Figure \ref{sampling} further shows that the policy-guided sampler generates pseudo-OOD samples clearly separated from ID clusters, while the Gaussian sampler adds isotropic noise around the midpoint of two randomly selected cluster centers, yielding less distinguishable samples. 

\subsubsection{Sensitivity Analysis of the Number of Prototypes.}
\begin{figure}[t]
\centering
\includegraphics[width=1.0\columnwidth]{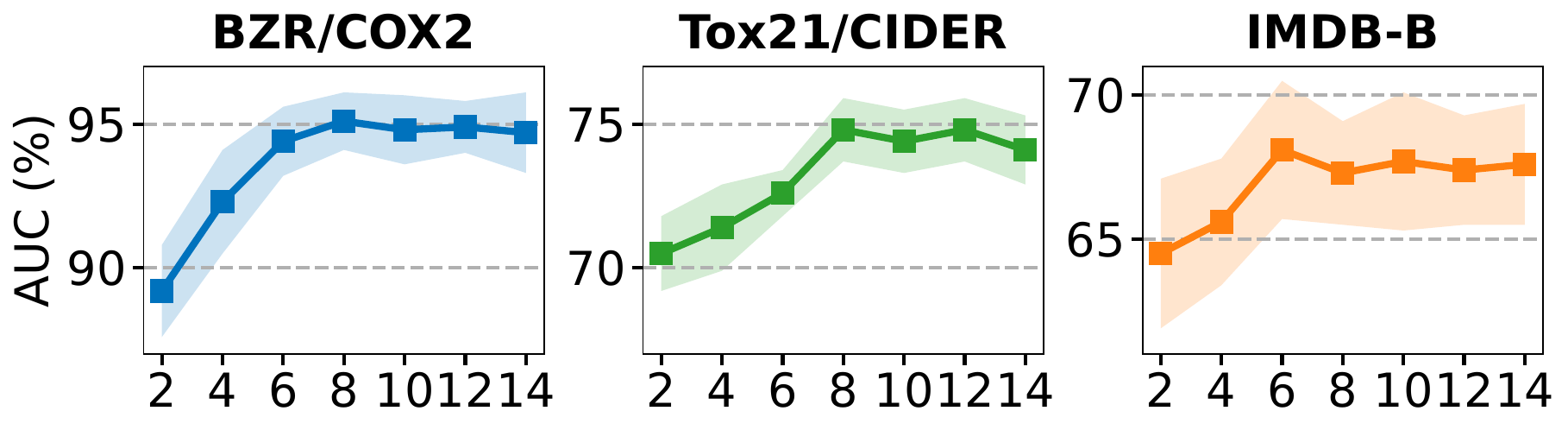} 
\caption{Sensitivity analysis of the number of prototypes $K$ on OOD detection performance across different datasets.}
\label{hyperparam_study}
\end{figure}

Figure \ref{hyperparam_study} illustrates the sensitivity of prototype count $K$ on 2 OOD dataset pairs and 1 anomaly detection dataset. A small $K$ performs poorly as too few clusters generate ineffective pseudo-OOD samples for regularization. Conversely, a sufficient K improves the representation of the ID data, which helps generate higher-quality outliers and enhances detection capability. This analysis confirms the importance of selecting an adequate number of prototypes to achieve optimal model performance.


\section{Conclusion}
We introduce PGOS, a framework for unsupervised graph OOD detection that learns an adaptive policy to replace static sampling. Our method first structures the latent space using Prototypical Graph Contrastive Learning, then deploys a policy-guided agent to explore low-density regions and generate informative pseudo-outliers. These pseudo-outliers are then used to train a robust OOD detection model. Experimental results demonstrate the effectiveness of our approach, with PGOS achieving strong performance and state-of-the-art results on the majority of OOD and anomaly detection benchmarks. Future work includes exploring more advanced reward mechanisms and extending our paradigm to other data modalities and security-related tasks.

\section*{Acknowledgements}
This work is supported in part by the NSFC under grant 62202164, the Shandong Provincial Natural Science Foundation under grant ZR2025MS1038, and the Young Scholars Program of Shandong University. Philip S. Yu is supported in part by the NSF under grants III-2106758 and POSE-2346158.

\bibliography{aaai2026}

\end{document}